\documentclass[10pt, a4paper]{article}

\usepackage[review]{lrec-coling2024} 

\title{Prediction of Translation Techniques for the Translation Process}

\name{Fan Zhou, Vincent Vandeghinste} 

\address{KU Leuven, KU Leuven \\
         Belgium, Belgium \\
         fan.zhou@student.kuleuven.be, vincent.vandeghinste@kuleuven.be}

\abstract{
Machine translation (MT) encompasses various methodologies aimed at enhancing translation accuracy. In contrast, the human-generated translation process relies on diverse translation techniques, which proves essential to ensuring both linguistic adequacy and fluency. This study proposes that these translation techniques can serve as a guide for optimizing machine translation further. However, it's imperative to automatically identify these techniques before they can effectively guide the translation process. The study distinguishes between two translation process scenarios: from-scratch translation and post-editing. For each scenario, a tailored set of experiments has been devised to predict the most suitable translation techniques. The results show that for from-scratch translation, the predictive accuracy reaches 82\%, while the post-editing process shows even greater promise, achieving an accuracy rate of 93\%.
 \\ \newline  }

\begin{document}

\maketitleabstract

\section{Introduction}

Neural machine translation (NMT) systems, such as Google Translate\footnote{\url{https://translate.google.com/}}, DeepL\footnote{\url{https://www.deepl.com/en/translator}}, and recent large language models like ChatGPT\footnote{\url{https://chat.openai.com/}} \citep{hendy2023good}, have made significant progress but still fall short of achieving human parity (\citealp{even1979polysystem}; \citealp{toury1991descriptive}; \citealp{baker1993corpus}; \citealp{baker1996corpus};  \citealp{baker2004corpus}; \citealp{dayrell2007quantitative};  \citealp{hassan2018achieving}; \citealp{wang2021progress}). Issues persist, such as word-for-word translation, false friends, ambiguity, information omission or addition, cultural insensitivity, etc. (\citealp{pușnei2023source}; \citealp{zhang2023disambiguated}; \citealp{bao2023finegrained}; \citealp{lee2023hate}; \citealp{hendy2023good}), resulting in low-quality translations that may not be easily understood.

\begin{figure*}
    \centering
    \includegraphics[width=1\textwidth]{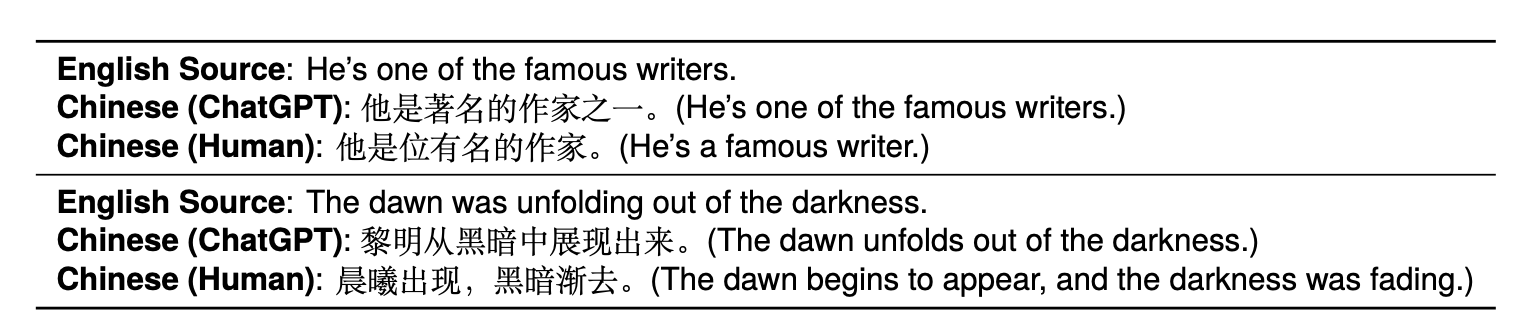}
    \caption{Comparison of English-Chinese Translation Examples: Machine Translation vs. Human Translation}
    \label{figure_chinese}
\end{figure*}

In the analysis of low-quality translations, most issues stem from inappropriate translation techniques. For example, in Figure \ref{figure_chinese}, both source sentences were translated using a literal or word-for-word technique with ChatGPT, resulting in either unnatural Chinese expressions or poor comprehension. In contrast, the first human translation employs the "reduction" translation technique to convey the meaning more concisely, while the second human translation splits the sentence into two clauses, using "modulation" for a more natural and concise translation.

Utilizing translation techniques is crucial for addressing translation problems, improving translation quality, and ensuring contextually appropriate translations. However, there are few existing studies explicitly suggesting such inference processing in current MT systems as to humans' decisions on translation techniques throughout the translation procedure (\citealp{stahlberg2020neural}; \citealp{dabre2020survey}; \citealp{wan2022challenges}; \citealp{ranathunga2023neural}; \citealp{klimova2023neural}).

These translation techniques can serve as a guide for human translation and, furthermore, for the generation of MT. They can also act as prompts for large language models to produce high-quality translations. In our investigation of both from-scratch translation and post-editing scenarios, our objective is to ascertain whether it is possible to predict the most appropriate translation techniques when provided with the source sentences and bad translations together with its source sentences. Our experiments have shown that pre-trained models, after being fine-tuned for both scenarios, can proficiently predict the most suitable translation techniques. This paves the way for future advancements in the field of machine translation generation.

\section{Related Work}

In the from-scratch translation process, manipulating the NMT architecture can help improve translation accuracy. For instance, structuring hidden states in the encoder as syntactic tree representations has shown performance improvement \cite{chen2017improved}. Another approach involves a semantic interface that connects pre-trained encoders and decoders, enabling separate pre-training with a shared language-independent space \cite{ren2021semface}. In the post-editing process, \citet{chatterjee2019findings} developed automatic post-editing (APE), which automatically corrects errors in the output of an MT system by learning from human corrections. APE can be viewed as a cross-lingual sequence-to-sequence task, which takes a source sentence and the corresponding MT output as inputs and generates the post-edited (PE) output \cite{lee2020cross}. Lee used source texts along with MT translation as input to be processed in the Transformer's encoder to generate the post-edited output, and then used its decoder to fine-tune PE. However, \citet{carl2017translation} explored the effects of cross-lingual syntactic and semantic distances on translation production times and found that non-literal translations pose challenges in both from-scratch translation and post-editing.\\
\indent Translation techniques assist in selecting appropriate translations, considering factors like naturalness and discourse coherence \cite{deng2017translation}. Non-literal translations reflect the diversity of human languages and are essential for preserving accuracy and fluency \cite{zhai2020building}. \citet{zhai2019recognizing} automated the classification of translation techniques for English-French pairs, but the number of aligned pairs for the experiment is limited. For the translation techniques used in NMT, most studies focus on idiom translation problems caused by literal translation. The fact is that most of the idioms are translated with non-literal translation techniques, but according to the evaluation of the language model \cite{dankers2022can}, Transformer's tendency to process idioms as compositional expressions contributes to poor literal translations of idioms. In order to make the idiom translated correctly, \citet{fadaee2018examining} retrieved sentences containing idioms and mixed them with non-idiom phrases as a training data set to train the NMT. Apart from idioms, some other literal translation problems are studied. For example, \citet{gamallo2021using} suggested that the large amount of literal translation causes transferring the constructions of the source language to the target language, and they proposed to use monolingual corpora instead of parallel ones with unsupervised translation, and this makes the hybrid SMT plus NMT system produce more non-literal translation output on passive voice-structured sentences. \citet{bacquelaine2023deepl} found word-by-word translation decreases in DeepL and Google Translate. The case study of \citet{raunak2023gpts} shows that there is a greater tendency towards non-literalness in GPT translations and GPT systems' ability to figuratively translate idioms.\\
\indent In this paper, we use English-Chinese pairs labeled with translation techniques to investigate whether pre-trained cross-lingual language models can be fine-tuned to accurately predict translation techniques so as to provide guidance for producing good translations in both from-scratch translation and PE/APE processes.

\section{Data}
Our study uses English-Chinese aligned pairs at the word and phrase level. Each aligned pair is labeled with a specific translation technique.

\subsection{UM Corpus}
The English-Chinese parallel sentences employed in our experiments are sourced from the UM-corpus \cite{tian2014corpus}. This corpus boasts manually verified sentence-level alignments. The parallel sentences we have selected for our study have been screened to ensure non-duplication and feature well-aligned sentence pairs. These specific pairs were chosen due to the presence of at least one individual aligned unit at the sub-sentence level (either a word or a phrase). This selection criterion allows for precise labeling with a specific translation technique, eliminating potential ambiguities where a single aligned unit might be associated with multiple labels.

\subsection{Data Preparation}
The data preparation process involves three phases: aligning parallel pairs, extracting features for each translation technique, and labeling the pairs.

\textbf{Alignment} In the alignment process, we work with parallel sentences, with a specific focus on sub-sentence units. English and Chinese sentences are tokenized using SpaCy\footnote{\url{https://spacy.io/}} and Jieba\footnote{\url{https://github.com/fxsjy/jieba}}, respectively. To ensure precise alignment of English-Chinese tokens, we use multiple toolkits, such as GIZA++ \cite{och2003systematic}, and cross-lingual models like mBERT \cite{devlin2018bert}, mBART \cite{liu2020multilingual}, and mT5 \cite{xue-etal-2021-mt5}. These models calculate cosine similarity and select the highest score for each word in the source language, significantly improving alignment accuracy. Furthermore, we employ custom algorithms to retrieve data that specifically pertains to translation technique features. To ensure data's high quality, we incorporate human annotations and leverage ChatGPT's alignment by prompts.

\textbf{Features of Translation Techniques} Our feature selection process adheres to the guidelines outlined in the "Annotation Guidelines of Translation Techniques for English-Chinese"\footnote{\url{https://yumingzhai.github.io/files/Annotation_guide_EN_ZH.pdf}}. These guidelines present a comprehensive typology of eleven translation techniques: literal translation, equivalence, transposition, modulation, modulation+transposition, particularization, generalization, figurative translation, lexical shift, explication, and reduction. Notably, we exclude "figurative translation" from this list for three reasons: 

\begin{itemize}
\item scarcity in the data;
\item elusive features for detection;
\item the fact that using the other ten techniques can also yield good translations.
\end{itemize}

From a linguistic perspective, we categorize these features into three types: 
\begin{itemize}
\item syntax-featured;
\item semantics-featured;
\item both syntax-featured and semantics-featured.
\end{itemize}

To collect data based on the features of each translation technique, we employ a range of methods. For translation techniques with syntactically related features, such as transposition and lexical shift, we employ SpaCy for part-of-speech tagging. For techniques associated with semantic features, such as equivalence, particularization, and generalization, we use Synonym\footnote{\url{https://github.com/chatopera/Synonyms}}'s word similarity, NLTK WordNet\footnote{\url{https://www.nltk.org/howto/wordnet.html}}'s hyponyms and hypernyms, along with prompt-based ChatGPT for initial processing. We also incorporate human checks to ensure data quality. Translation techniques that involve both syntactic and semantic features, like modulation and modulation+transposition, are handled by combining the aforementioned methods to collect the data.

\begin{table}
\small
\centering
\begin{tabular}{llllllll}
\toprule
\textbf{Task\qquad \qquad \qquad Softmax}& \textbf{Train} & \textbf{Dev} & \textbf{Test} \\
\midrule
Architecture 1 \qquad \qquad  10 & 69828 & 7759 & 8621 \\
Architecture 2 \qquad \qquad  10 & 69828 & 7759 & 8621 \\
\midrule
Architecture 3\textsubscript{subtask1} \quad \ \   3 & 73958 & 8218 & 9131\\ 
\qquad \qquad \qquad  \textsubscript{subtask2} \quad \ \   9 & 54303 & 6034 & 6890 \\
\midrule
Architecture 4 \qquad \qquad  10 & 73958 & 8218 & 9131 \\
\bottomrule
\end{tabular}
\caption{\label{table_data_split} Data Overview}
\end{table}

\textbf{Labeling} In the labeling process based on features, we assign each aligned pair a single, unequivocal translation technique.

\subsection{Data Summary}

We have amassed a dataset of more than 100,000 data pairs. Each data pair consists of a source sentence, a target sentence, an aligned word or phrase (a source unit and a target unit), and a corresponding translation technique label. For a comprehensive view of how the data is employed for specific tasks, please refer to the details provided in Table \ref{table_data_split}. We make both the collected data and the codes for data retrieval and experiments available\footnote{\url{https://anonymous.4open.science/r/translation_technique-436D/}}.

\begin{figure*}
    \centering
    \includegraphics[width=1\textwidth]{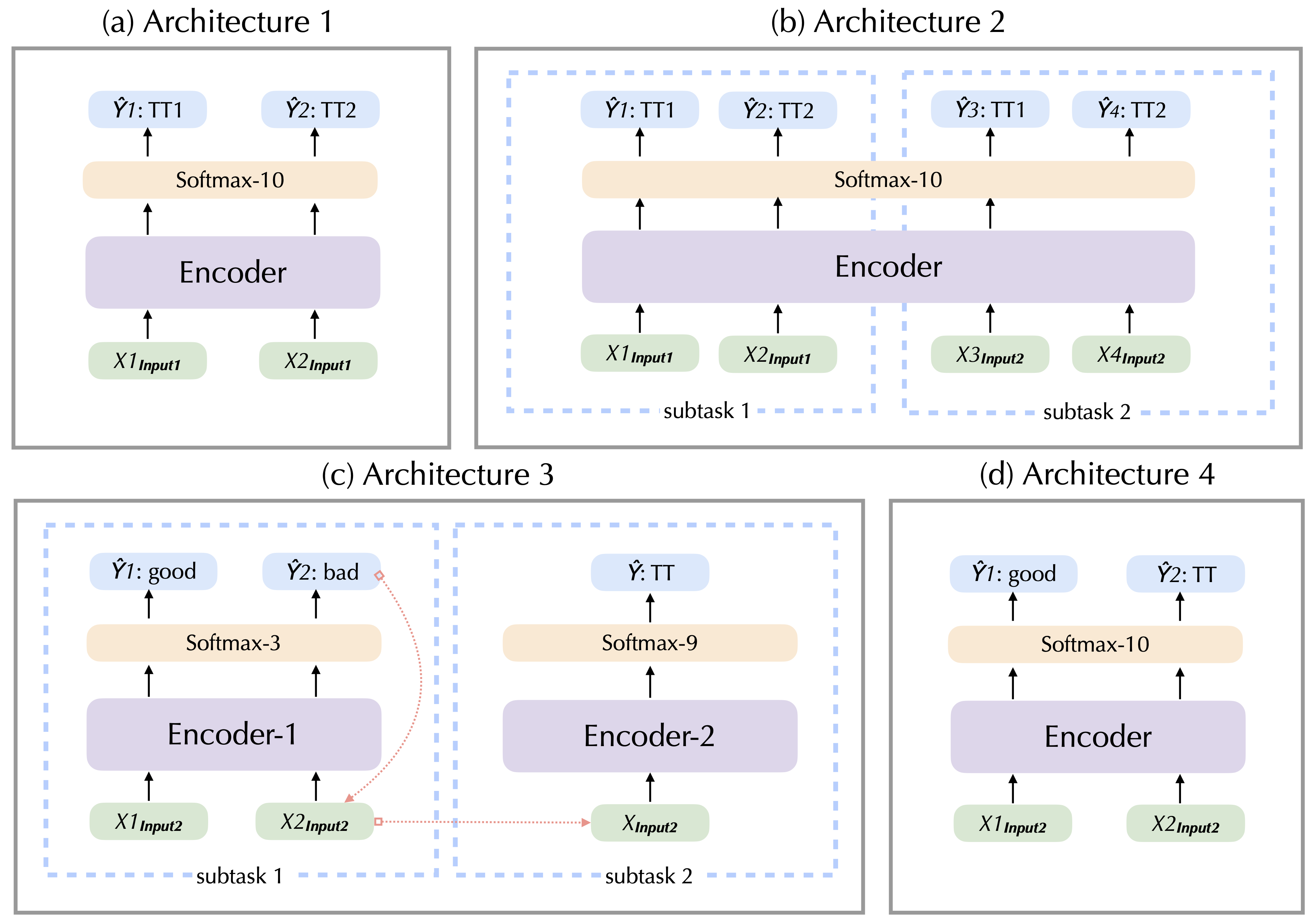}
    \caption{Experiments feature four distinct architectures: (a) and (b) are designated for from-scratch translation, while (c) and (d) are tailored for post-editing tasks. The input data is structured in two formats: Input1 and Input2, with comprehensive specifics available in Figure \ref{figure_chinese2}. 'n' of the Softmax-n layer corresponds to the number of categories. 'TT' signifies a specific translation technique.}
    \label{figure1}
\end{figure*}

\begin{figure*}
    \centering
    \includegraphics[width=1\textwidth]{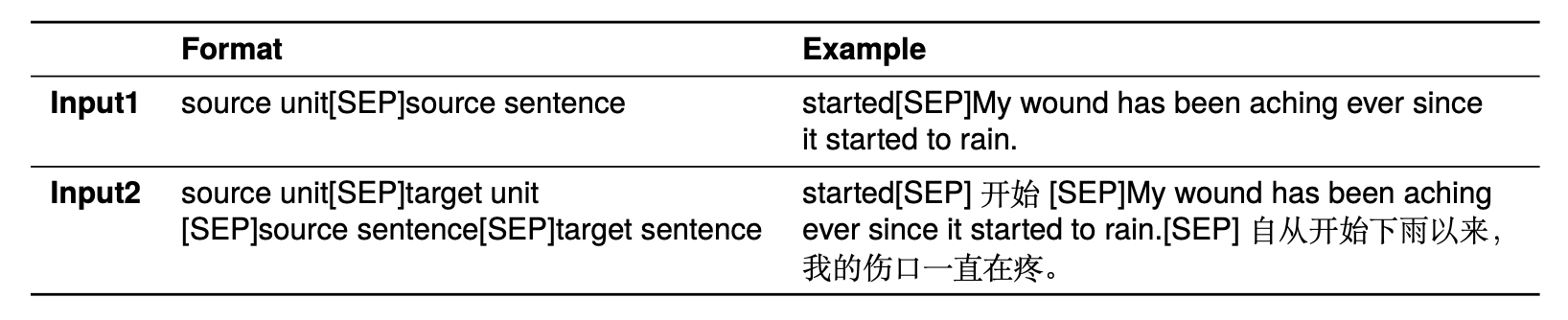}
    \caption{Data Input Formats}
    \label{figure_chinese2}
\end{figure*}

\begin{table*}
\small
\centering
\begin{tabular}{@{}clc@{}}
\toprule
&\textbf{Model} & \textbf{Architecture 1} \\
\midrule
&BERT & 0.802 \\
&BART & 0.774 \\
&T5 & \textbf{0.811} \\
\bottomrule
\end{tabular}
\qquad
\begin{tabular}{@{}clcccc@{}}
\toprule
&\textbf{Model} & \textbf{Architecture 2}& \multicolumn{2}{c}{\textbf{Architecture 3}} & \textbf{Architecture 4} \\
\midrule
&mBERT&0.808 & 0.912\textsubscript{subtask1} & \textbf{0.929}\textsubscript{subtask2} & 0.901 \\
&mBART&0.813 & 0.911\textsubscript{subtask1} & 0.922\textsubscript{subtask2} & 0.907 \\
&mT5 & \textbf{0.819} & \textbf{0.915}\textsubscript{subtask1} & 0.920\textsubscript{subtask2} & \textbf{0.914} \\
\bottomrule
\end{tabular}
\caption{\label{model_overall_performance} Overall performance by different models in four architectures, measured by accuracy.}
\end{table*}

\begin{table*}
\small
\centering
\begin{tabular}{@{}lllllllllllll@{}}
\toprule
 \textbf{Architecture 1} & \textbf{LIT} & \textbf{TRA} & \textbf{EXP} & \textbf{RED} & \textbf{LEX} & \textbf{EQU} & \textbf{GEN} & \textbf{MOD} & \textbf{PAR} & \textbf{MOT} \\
\midrule
BERT-Base & 0.725 & 0.672 & 1.0 & \textbf{0.923} & 0.784 & \textbf{0.900} & 0.728 & \textbf{0.773} & 0.775 & 0.579 \\
 BART-Large & 0.681 & 0.657 & 1.0 & 0.901 & 0.736 & 0.891 & 0.714 & 0.748 & 0.746 & 0.540 \\
T5-Large & \textbf{0.741} & \textbf{0.699} & 1.0 & 0.920 & \textbf{0.804$ \dag  $} & \textbf{0.900} & \textbf{0.751} & 0.764 & \textbf{0.782} & \textbf{0.623$ \dag  $} \\
\midrule
\midrule
 \textbf{Architecture 2} & \textbf{LIT} & \textbf{TRA} & \textbf{EXP} & \textbf{RED} & \textbf{LEX} & \textbf{EQU} & \textbf{GEN} & \textbf{MOD} & \textbf{PAR} & \textbf{MOT} \\
\midrule
mBERT & 0.740 & 0.710 & 1.0 & \textbf{0.936$ \dag  $} & 0.794 & \textbf{0.903$ \dag  $} & 0.728 & 0.761 & 0.773 & 0.585 \\
mBART-Large & 0.741 & 0.708 & 1.0 & 0.929 & 0.793 & 0.901 & \textbf{0.760$ \dag  $} & 0.769 & \textbf{0.806$ \dag  $} & \textbf{0.605} \\
mT5-Large & \textbf{0.762 $ \dag  $} & \textbf{0.727$ \dag  $} & 1.0 & 0.932 & \textbf{0.799} & 0.890 & 0.753 & \textbf{0.781$ \dag  $} & 0.793 & \textbf{0.605} \\
\midrule\midrule
\textbf{Architecture 3} & \textbf{BAD} & \textbf{TRA} & \textbf{EXP} & \textbf{RED} & \textbf{LEX} & \textbf{EQU} & \textbf{GEN} & \textbf{MOD} & \textbf{PAR} & \textbf{MOT} \\
\midrule
mBERT & 0.905 & \textbf{0.962$ \ast  $} & 1.0 & 0.984 & 0.979 & \textbf{0.905$ \ast  $} & \textbf{0.850$ \ast  $} & 0.922 & \textbf{0.880$ \ast  $} & \textbf{0.752$ \ast  $} \\
mBART-Large & 0.907 & 0.957 & 1.0 & 0.982 & \textbf{0.981$ \ast  $} & 0.888 & 0.840 & 0.921 & 0.868 & 0.720 \\
mT5-Large & \textbf{0.916} & 0.938 & 1.0 & \textbf{0.986$ \ast  $} & 0.979 & \textbf{0.905$ \ast  $} & 0.847 & \textbf{0.925$ \ast  $} & 0.854 & 0.700 \\
\midrule\midrule
 \textbf{Architecture 4} & \textbf{GOOD} & \textbf{TRA} & \textbf{EXP} & \textbf{RED} & \textbf{LEX} & \textbf{EQU} & \textbf{GEN} & \textbf{MOD} & \textbf{PAR} & \textbf{MOT} \\
\midrule
mBERT & 0.938 & 0.904 & 1.0 & \textbf{0.980} & 0.938 & 0.832 & 0.692 & 0.726 & 0.705 & 0.525 \\
mBART-Large & 0.943 & \textbf{0.919} & 1.0 & 0.961 & 0.902 & 0.792 & 0.706 & 0.686 & \textbf{0.736} & 0.570 \\
mT5-Large & \textbf{0.954$ \ast  $} & 0.893 & 1.0 & 0.974 & \textbf{0.955} & \textbf{0.839} & \textbf{0.712} & \textbf{0.742} & 0.710 & \textbf{0.604} \\
\bottomrule
\end{tabular}
\caption{\label{model_each_translation techniques}The models' predictions encompass various translation techniques, including LIT (literal translation), TRA (transposition), EXP (unaligned explicitation), RED (unaligned reduction), EQU (equivalence), GEN (generalization), MOD (modulation), PAR (particularization), MOT (modulation+transposition), BAD (bad translation), and GOOD (good translation). These predictions are assessed using the F1-score as the performance metric.}
\end{table*}

\section{Experiments}
\label{marker0}

There are four experiments involving two architectures of from-scratch translation (\ref{marker1} and \ref{marker2}) and two architectures of PE/APE (\ref{marker3} and \ref{marker4}). Since the experiments do not involve translation generation, our focus is solely on the encoder part of NMT architectures.

The four experiments, each employing different architectures, are depicted in Figure \ref{figure1}. The data overview for each architecture can be seen in Table \ref{table_data_split}. Different architectures utilize distinct data formats as input, encompassing two input formats, denoted as Input1 and Input2 in Figure \ref{figure_chinese2}.

\subsection{From-scratch Translation}
The sole data available for from-scratch translation consists of texts in the source language. In preparation for the from-scratch translation generation, we employ language models to explore the possibility of predicting appropriate translation techniques. This predictive analysis is carried out as part of the pre-translation process.

We employ two distinct architectures for this prediction task. One architecture is trained exclusively with source language data, while the other incorporates both source and target language data, with the target data serving as supplementary information to train the model. The objective of these two experiments is to forecast both literal and non-literal translation techniques for the source language.

\subsubsection{Architecture 1} 
\label{marker1}
We adopt the architecture showcased in Figure \ref{figure1} (a). In contrast to the approach employed by \citet{zhai2020building}, where they localized aligned pairs and extracted the last hidden state based on token positions, our methodology takes a different path. Our model's input combines the source part of the aligned pair with the source sentence by inserting the special token \texttt{[SEP]} (end-of-sentence token) between them, referred to as Input1 (as detailed in Figure \ref{figure_chinese2}). This format effectively unifies both parts into a single input, allowing us to consider how the sole source part within the source sentence interacts with the entire context, resembling the human translation process.

For fine-tuning, we employ encoder models, which encompass BERT-base, BART-large's encoder, and T5-large's encoder. These models vary in terms of structure and the number of trainable parameters (110M, 410M, and 770M parameters, respectively). The fine-tuning process stops when accuracy reaches its peak on the validation dataset. We utilize the Adam optimizer with an initial learning rate of 1e-5, and the learning rate is dynamically adjusted. Additionally, dropout or L2 regularization is dynamically incorporated in different training epochs. Typically, fine-tuning converges at the sixth epoch.

\subsubsection{Architecture 2}
\label{marker2}

To improve the accuracy of classifying and predicting translation techniques, we introduce a multi-task architecture, which is illustrated in Figure \ref{figure1} (b). This architecture comprises two subtasks. Subtask 1 is focused on predicting translation techniques based on data from the source language, and its input follows the Input1 format, as outlined in Figure \ref{figure_chinese2}. On the other hand, Subtask 2 is designed to identify patterns associated with specific translation techniques, utilizing Input2 (Figure \ref{figure_chinese2}) as its input. Subtask 2 encodes data from both the source and target languages, providing the model with insights into how translation techniques are utilized. Both subtasks share the same encoder, employing identical architecture and parameters.

Given that the primary goal of the experiments related to the from-scratch translation process is to use source language data exclusively for predicting the most suitable translation technique for each unit within a source language sentence, subtask 1 holds the principal role, with subtask 2 serving as a supportive auxiliary task. Consequently, during the back-propagation process, we calculate a weighted loss summation for the optimization strategy, as outlined in Formula (\ref{formula1}). The weight configuration, with $\alpha$ set at 0.8 for subtask 1 loss and $\beta$ at 0.2 for subtask 2 loss, yields optimal results. This weight distribution emphasizes that the second task is the primary one and should carry more weight during the encoder training.

\begin{equation}
L = \alpha L1 + \beta L2
\label{formula1}
\end{equation}

In this setup, we evaluate cross-lingual encoder models, including mBERT, mBART-large, and mT5-large, which have 110M, 410M and 770M parameters, aligning with parameter numbers of models used in Architecture 1. The training process stops when the best validation accuracy is achieved.

\begin{figure*}
    \centering
    \includegraphics[width=1\textwidth]{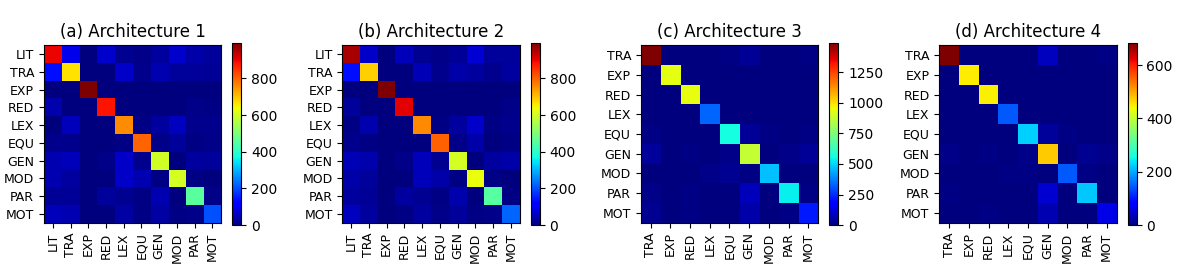}
    \caption{Translation techniques' prediction heatmap. The statistical number of each translation technique is averaged from 3 models in each architecture. Non-diagonal X-axis represents false positive and non-diagonal Y-axis represents false negative.}
    \label{fig:prediction_error_heatmap}
\end{figure*}

\subsection{PE/APE}

Our investigation aims to determine whether models can accurately predict suitable translation techniques for poor translations in the PE/APE process. In practice, post-editors frequently need to differentiate between good and poor translations prior to the PE/APE process. As a result, our experiments unfold in two steps:

\begin{enumerate}
\item \texttt{Identifying poor translations from the provided dataset;}
\item \texttt{Predicting appropriate translation techniques for these poor translations.}
\end{enumerate}

We conduct two distinct experiments for the prediction task that precedes the PE/APE process. The first architecture dissects these two steps, while the second integrates them into a single process, where the system distinguishes poor translations and simultaneously predicts the required translation techniques.

Bad translations can arise from various factors, with one of the most common causes being the application of the literal translation technique. This translation technique often lead to translations that are rigid or entirely incorrect. Therefore, in these two experiments, we introduce "bad literal translations" as the exclusive category of bad translations, which are generated from the existing good translations by replacing good non-literal translations with bad literal translation.

\subsubsection{Architecture 3}   
\label{marker3}
In this architecture, we employ two encoders, and the structure, depicted in Figure \ref{figure1} (c), consists of two consecutive subtasks. Both encoders utilize the Input2 format, as specified in Figure \ref{figure_chinese2}.

In subtask 1, we employ Encoder-1 to segregate poor translations from the good translations which encompass both good literal and good non-literal translations. After this filtering process, good translations remain, while the poor ones are directed to subtask 2.

Within subtask 2, Encoder-2 is utilized to assign suitable translation techniques to the identified poor translations. Encoder-2 is trained using data that includes all poor translations, specifically, rigid, poor literal translations. These bad literal translations are fed into Encoder-2, and its predictions include one of nine non-literal translation techniques. This setup enables us to rectify bad literal translations by applying non-literal translation techniques during the PE/APE phase.

To facilitate PE, post-editors initially need to identify bad translations within the source texts and their corresponding translations. We kickstart this process by fine-tuning a pre-trained model for this purpose. The encoders used in the two subtasks of this experiment are of the same type of cross-lingual language models, including mBERT, mBART-large's encoder, and mT5-large's encoder. Encoder-1 converges at the fourth epoch of the training process, while Encoder-2 reaches convergence at the fifth epoch.

\subsubsection{Architecture 4}
\label{marker4}
The architecture utilized in this experiment, illustrated in Figure \ref{figure1} (d), consolidates the two-step process of Architecture 3 into a single step, featuring just one encoder. All data, formatted as Input2 (as shown in Figure \ref{figure_chinese2}), which includes both good and bad translations, is fed into this encoder. The bad translations receive appropriate translation techniques, while the output for the good translations is labeled as "good." For this experiment, we employed the encoders of mBERT, mBART-large, and mT5. The encoder reaches convergence at the fifth epoch of the training process.

\section{Results}

Table \ref{model_overall_performance} showcases the overall performance of the models, as assessed by their accuracy. Table \ref{model_each_translation techniques} presents the models' predictions, measured by the F1-score, for different translation techniques. Moreover, Figure \ref{fig:prediction_error_heatmap} provides an illustration of the likelihood of incorrect predictions for the translation techniques.

In both Architectures 1 and 2, the overall accuracy hovers at approximately 81\%. However, the latter displays slightly higher average prediction accuracy. Models fed with the even distribution of input data results in an uneven prediction output. Translation techniques with syntax features achieve prediction  F1-score exceeding 80\%, surpassing those with semantic features which range between 70-80\%. Both architectures face challenges when predicting modulation+transposition, achieving a F1-score of only around 61\%.

In Architecture 3, the accuracy for subtask 1, which involves models distinguishing bad literal translations from good translations, surpasses 90\%. The overall accuracy reaches as high as 92\% in subtask 2. When predicting appropriate translation technique labels for bad translations, most syntactic-related translation techniques (over 94\%) and certain semantics-related translation techniques (equivalence and modulation at 90\%, generalization and particularization at 85\%) achieve high F1-scores in their predictions. However, the modulation+transposition technique only reaches the best F1-score of 75.2\% when using mBERT.

In Architecture 4, the models' performance achieves an accuracy above 90\%. Specifically, they can identify "good translations" with an impressive 95\% F1-score when utilizing mT5. When predicting translation techniques for bad translations, syntax-related techniques achieve over 90\% on F1-score, surpassing semantics-related techniques. Notably, generalization (71.2\%, best F1-score achieved with mT5), particularization (73.6\%, best F1-score achieved with mBART), modulation (74.2\%, best F1-score achieved with mT5), and modulation+transposition (60.4\%, best F1-score achieved with mT5) exhibit comparatively lower F1-scores.

Both Architectures 3 and 4 achieve high accuracy. However, they differ in structures and dataset used. Whether the difference in prediction performance is due to the architectural design or the dataset composition remains to be explored.

\section{Conclusions}

Using translation techniques is useful in the human translation process to ensure translation accuracy and fluency, yet it remains an under-discussed aspect in current neural machine translation. In this study, we explore the feasibility of predicting translation techniques for different words and phrases within their contexts. This prediction not only aids in human translation but also serves as a means to provide prompts to large language models for the generation of precise translations.

However, it's important to note that our current work primarily focuses on the phases preceding the from-scratch translation and PE/APE process, without directly incorporating translation techniques into NMT systems for translation generation. Our future research will focus on enhancing the decoder component to achieve optimal translations, guided by translation techniques. Additionally, we acknowledge the challenges associated with obtaining substantial dataset for experiments, considering the scarcity of open-sourced parallel sentence corpora with aligned lexical or phrasal units as well as inaccurate alignment by existing tools. Our forthcoming efforts will also focus on the automation of sub-sentence parallel unit alignment to streamline research and reduce the burden on human resources.

\nocite{*}
\section{Bibliographical References}\label{sec:reference}
\bibliographystyle{lrec-coling2024-natbib}

\bibliography{lrec-coling2024-example}

\label{lr:ref}
\bibliographystylelanguageresource{lrec-coling2024-natbib}
\bibliographylanguageresource{languageresource}

\begin{table*}[h!]
\centering
\begin{tabular}{p{5cm}l}
\toprule
\textbf{Translation Techniques} & \textbf{Features}\\
\midrule
Literal translation & Verb \\
                   & Noun \\
                   & Noun \\
                   & Adjectives\\
                   & Adverbs\\
                   & Pronouns\\
                   & Prepositions\\
                   & Negative expressions\\
                   & Passive voice\\
                   & Measure units\\
\midrule
Equivalence & Proverbs, idioms, and fixed expressions \\
                       & Cultural equivalence\\
                       & Measurement units\\
                       & Abbreviation\\
\midrule
Transposition & From verb to other parts of speech (POS) \\
                         & From noun to other POS\\
                         & From adjective to other POS\\
                         & From adverb to other POS\\
                         & From pronoun to other POS\\
                         & From preposition to other POS\\
\midrule
Modulation & Metonymical modulations \\
                      & Change the point of view\\
                      & Change between passive voice and active voice\\
                      & Affirmative form to negative form, the negation of the opposite\\
                      & The subject becomes the object\\
                      & Obligatory syntactic change but no change in meaning\\
                      & Circumvent translation difficulties, achieve expression naturalness\\
                      & Slight meaning change in lexical level according to the context\\
\midrule
Modulation + Transposition & Prepositions \\
                                      & Nouns\\
\midrule
Particularization & Specify the meaning of a word \\
                             & Translate a pronoun by the thing(s) it refers to\\
\midrule
Generalization & Removal of a metaphorical image \\
                          & Pronoun to translate the thing(s) that it references\\
\midrule
Lexical Shift & Change verbal tense \\
                         & Differences between plural and singular form\\
                         & Remove the passive voice\\
\midrule
Explicitation & Resumptive anaphora \\
                         & Add Chinese-specific words\\
                         & Add logical connectives\\
\midrule
Reduction & Removal of preposition \\
                     & Removal of determiner\\
                     & Removal of noun\\
                     & Removal of pronoun\\
                     & Removal of copula\\
                     & Removal of anticipatory "it"\\
\bottomrule 
\end{tabular}
\caption{Features of Translation Techniques}
\label{features_translation_techniques}
\end{table*}

\end{document}